# MatriVasha: A Multipurpose Comprehensive Database for Bangla Handwritten Compound Characters

This Report Presented in Partial Fulfillment of the Requirements for the Degree of Bachelor of Science in Computer Science, and Engineering

**Supervised by:**

Dr. Syed Akhter Hossain

Professor and Head

Department of Computer Science and Engineering

Daffodil International University

E-mail: aktarhossain@daffodilvarsity.edu.bd

A K M Shahariar Azad Rabby

Lecturer,

Department of Computer Science and Engineering

Daffodil International University

E-mail: shahariar.cse@diu.edu.bd, shahariarrabby@gmail.com

**Submitted by:**

Jannatul Ferdous

ID:

Department of CSE

Daffodil International University

Email: jannatul15-7287@diu.edu.bd

Suvrajit Karmaker

ID:

Department of CSE

Daffodil International University

Email: suvrajit15-6944@diu.edu.bd

# ABSTRACT

At present, recognition of the Bangla handwriting compound character has been an essential issue for many years. In recent years there have been application-based researches in machine learning, and deep learning, which is gained interest, and most notably is handwriting recognition because it has a tremendous application such as Bangla OCR. MatriVasha, the project which can recognize Bangla, handwritten several compound characters. Currently, compound character recognition is an important topic due to its variant application, and helps to create old forms, and information digitization with reliability. But unfortunately, there is a lack of a comprehensive dataset that can categorize all types of Bangla compound characters. MatriVasha is an attempt to align compound character, and it's challenging because each person has a unique style of writing shapes. After all, MatriVasha has proposed a dataset that intends to recognize Bangla 120(one hundred twenty) compound characters that consist of 2552(two thousand five hundred fifty-two) isolated handwritten characters written unique writers which were collected from within Bangladesh. This dataset faced problems in terms of the district, age, and gender-based written related research because the samples were collected that includes a verity of the district, age group, and the equal number of males, and females. As of now, our proposed dataset is so far the most extensive dataset for Bangla compound characters. It is intended to frame the acknowledgment technique for handwritten Bangla compound character. In the future, this dataset will be made publicly available to help to widen the research.

# TABLE OF CONTENTS

## List of Figures

# List of Tables

# CHAPTER 1

# Introduction

## 1.1 Introduction

Researchers always want to research different types of applications, and many countries are researching to introduce their handwritten recognition, text, and characters. Currently, automatic handwriting compound character recognition is one of the most critical research areas for various applications like O.C.R. which helps to identify the character from the images, and that's process consumes the time, and also research this application last few years, handwritten hardcopy information covert into a digital file format which is more reliable., and that's a result this system is comfortable to handle the document. But till now Bangla has no complete dataset that contains all of these characters, and compound words, and that's the result a large number of older documents are written by hand. This handwritten document is not added to the digital file, and this is the challenge of trying to convert the documents by typing the manual and following the traditional method for copying them. Most of the time we need to convert the digital file from the handwritten file. But if we do it manually, then this works will waste our colossal time. But we can do it using O.C.R. Using O.C.R., and we can easily convert our handwritten file to digital file. It saves our enormous time. But there is no reliable dataset for Bangla character recognition. Many works have been done, but those are concentrated for digit [1] or basic characters [2] or a few numbers of compound characters [3]. But works with handwritten characters are complicated because of their different shapes and styles for a reason for the different types of people., and the format of the Bengali letter is complicated due to its alignment. Many of them are similar in addition to the complementary compound letters of the other basic characters.

The basic language of one language is different from other languages like Latin scripts are different from Bengali because Bengali comes from the Sanskrit script. On the other hand, there is also every country's language that is different from other languages. Bangla compound characters this time are many unknown students and people. They do not know how to create a sentence using compound characters. So MatriVasha dataset help to known the compound character. On the other hand, many countries are also starting their research for their languages, and they got interested in this topic. In Bangla language, there are have 50 basic characters, ten numerical digits, more than 200 compound characters, and ten modifiers.

**1.2 Motivation**

Bengali or Bangla language is native to Bengal. Bengali speaks their mother tongue, which they feel comfortable, and they can also represent their thoughts. Bangla is the state language of Bangladesh and one of 18 languages listed in the Indian constitution. Bangla speaker's number about more than 230 million today, Bangla is the seventh language after English, Chinese, Hindi-Urdu, Arabic, Spanish, and Portuguese. February 21st is announced as the international mother language day by UNESCO to respect the language martyrs for the Bangla language in Bangladesh in 1952. People remember this day with great respect. We achieved our mother languages for our martyrs., and there are belongs to a different type of character. Bangla handwritten compound character is one of them. Bangla handwritten compound character recognition plays a vital role in helping those people in various purposes such as Bangla character recognition, automatic postal code identification, extracting data from hardcopy forms, intuitive I.D. card reading, automatic reading of bank cheques, and digitalization of documents, etc. On the other hand, it also helps to write or reading purposes as follows:

1. The Bengali language has been officially recognized by the constitution of India and the second most beautiful language in the world.
2. Bangla Literature contribution is not less than another language. Rabindranath Tagore was awarded the Nobel Prize for Gitanjali, which was written in Bangla.
3. Bangla language gets recognition in international mother language day, and also respect our martyrs to bring our freedom.
4. Bangla's handwritten compound character recognition is a robust model for Bangla language.
5. Kazi Nazrul Islam also awarded many prizes for poems, stories, etc. which was written in Bangla.

### 1.3 Rationale of the Study

Bangla handwritten compound character recognition is an interesting issue because of handwritten convert into images. On the other hand, Bangla's handwritten compound character not only uses in recognition but also uses to picture to text, text to speech work, and also forensic analysis to using handwriting. To collecting compound characters, it maybe helps to get much information. We collected data from various ages people and labeled as their given information like age, hometown, gender, education, and made this raw data into image data. Bangla handwritten compound character recognition is contained 120 various types of compound character which is a help to people to understand the compound character, and also known to the many unknown compound character, and its support to their educational purpose. Because many students do not identify some compound character. So it also helps to their educational purposes, and they are also giving their opinion after writing the different types of compound characters.

### 1.4 Research Questions

1. Do we need a lot of datasets to detect handwritten letters?
2. Why do we collect a large number of dataset from different types of people?
3. Why did we label the MatriVasha dataset as a people's gender, district, age, and education?
4. Can we pre-process the raw data to be used for the deep learning approaches?
5. Can the Convolution Neural Network Classifier algorithm be used on the preprocessed data?
6. How do they feel when they write compound characters?

### 1.5 Expected Outcome

1. A dataset for Bangla handwritten compound character dataset, and recognition.
2. MatriVasha which is the largest dataset for the compound character recognition.

3. Bangla's handwritten compound character is also research in both quality and quantity.
4. Our target is to build the Bangla language upward in the world.
5. Data processing method for handwritten data for any language.
6. People show they're interested in the complex characters.

### 1.6 Report Layout

There are five chapters in this research paper. They are Introduction, Literature Review, Proposed Research, Results, and Discussion, Conclusion, and Future.

**Chapter one:** Introduction; Objective, Motivation, Expected Outcome, Report layout.

**Chapter two:** Literature Review; Sensibility Analysis, Related works, challenges.

**Chapter three:** Proposed Research; Research Methodology, Data Collection, Data

Processing, Flow Model, Experimental layout.

**Chapter four:** Results, and Discussion; Experimental Result, Discussion.

**Chapter five:** Conclusion, and Future; Conclusion, Future Research.

# CHAPTER 2

# Background

## 2.1 Introduction

Bangla's handwritten compound character is a highly interesting topic in the academic and commercial research field. Because of, research field are mostly used in Bangla compound character. Many types of research held on Bangla handwriting and other languages like English, Arabic, Hindi, Chinese, etc. But the less research for different types of compound characters. So, the background chapter explanation of related work, a summary of this research, the scope of the problem, and lastly, challenges of this research.

## 2.2 Related Work

According to past studies have included several tasks for recognizing handwritten character different languages such as Latin [4], China [5], and Japanese [6] have had great success. On the other hand, a few works are available for Bangla handwritten basic character, digit, and compound character recognition. Some literature has been made on Bangla character recognition in the past years as "A complete printed Bangla O.C.R. system"[7], "On the development of an optical character recognition (O.C.R.) system for printed Bangla script [8]. There are also belong to a few types of research on handwritten Bangla numerical recognition that reaches to the desired recognition accuracy. "Automatic recognition of unconstrained offline Bangla handwritten numerals"[9], "A system towards Indian postal automation"[10] which according to pal et al. has conducted some exploration work for handwriting recognition of Bangla Characters. There are also schemes which names are "Touching numeral segmentation using water reservoir concept"[11]. These schemes are mainly based on extracted features from a concept called the water reservoir. Besides, there are several Bangla handwriting present received character recognition, and achieved pretty success. Halima Begum et al., “Recognition of Handwritten Bangla Characters using Gabor Filter, and Artificial Neural Network” [12] works with their dataset that was collected from 95 volunteers, and the proposed model achieves feature extraction, and without surrounding feature extraction 68.9%, and 79.4% of recognition rate respectively.

“Recognition of Handwritten Bangla Basic Character, and Digit Using Convex Hall Basic Feature” [13] achieve accuracy for Bangla characters 76.86%, and Bangla numerals 99.45%. “Bangla Handwritten Character Recognition using Convolutional Neural Network” achieved 85.36% test accuracy using their dataset. In “Handwritten Bangla Basic, and Compound character recognition using M.L.P., and SVM classifier” [14], the handwritten Bangla basic and compound character recognition using M.L.P., and SVM classifier has been proposed. They achieved around 79.73%, and 80.9% of recognition rate, respectively."Ekush: A Multipurpose and Multitype Comprehensive Database for Online Off-line Bangla Handwritten Characters" achieved 97.73% for the Ekush dataset.

There are also four open-access datasets available for Bangla characters which play a vital rule in recognizing handwritten characters dataset. These are the BanglaLekha-Isolated [15], the CMATERdb [16], and the I.S.I. [17], the Ekush [18]. BanglaLekha–Isolated dataset consists of a total of 166,105 squared images (while preserving the aspect ratio of the characters), each containing a sample of one of 84 different Bangla characters, which has three categories such as ten numeral digits, 50 basic characters, and 24 compound characters. Two other datasets CMATERdb also has three different categories for basic characters, numerals, and compound characters, and the I.S.I. The dataset has two different datasets for basic characters and numerals. Finally, Ekuash also has four different categories for modifiers, basic characters, numerals, and compound characters.

### 2.3 Research Summary

Our aim to make a model that can recognize Bangla handwritten compound character recognition using the convolutional neural network, which is training by MatriVasha, and other Bangla compound character datasets. Convolutional neural network (CNN) has revealed new opportunities in the field pattern recognition for categorization, which helps numerous researchers apply their sophisticated system to the solution. The CNN algorithm was complex; that's why it's not used thoroughly, and the structure was first proposed by Fukushima et al. in 1980 [18]. On the other hand, In the 1990s, Lacan et al. implemented a gradient-based learning algorithm on CNN and achieved successful results [19]. Researchers now work on CNN and made good results in recognition. A few years ago, Ciaran et al. [20] applied digits, alpha numbers, traffic signs, and other object classes. A CNN consists of an input, and an output layer, as well as multiple hidden layers. In deep learning, a convolutional neural network is a class of deep neural networks, commonly use in visual image analysis. CNN help to many researchers to help to get proper results. In recent times CNN is more accessible with researchers because they can get relevant results, but a few years ago, CNN is not favored with the people.

Table 2.1: Comparison of the number of images with our proposed dataset.

| Dataset Name | Unique Compound Characters | Total Compound Characters |
| --- | --- | --- |
| Ekush[18] | 50 | 150,840 |
| CMATERdb[16] | 160 | 42,248 |
| BanglaLekha-Isolated[15] | 24 | 47,407 |
| I.S.I. [17] | none | none |
| MatriVasha | 120 | 306,240 |

In table 2.1, where we can see the comparison between MatriVasha dataset, and the other four accessible sources of Bangla handwriting related datasets (BanglaLekha-Isolated dataset, CMATERdb, Ekush, and the I.S.I. handwriting dataset). The MatriVasha dataset consists of 30624 images that contain 120 unique compound characters and the most extensive dataset for Bangla compound characters where belong to different types of complex figures.

The proposed method is shown recognition accuracy of 94.49% for MatriVasha datasets.

## 2.4 Scope of the problem

- Recognize Bangla's handwritten documents, which include complex characters.
- Identify Bangla compound handwriting from photos.
- Elevate Bangla's picture to text, and picture to speech work.
- To help forensic analysis from handwriting.
- Handwriting analysis to public personal information.

- Compound character analysis also helps to educational background.

### 2.5 Challenges

The following are the challenges:

- Difficult to collect data: our target to collect data from 2500 unique people, which was divided equally, two groups, male and female. So, it was challenging to collect data similarly.
- We face the problem of taking permission to different schools and colleges.
- Many students are facing the problem of recognized compound characters. But they know many unknown compound characters.
- Preprocessing huge, and massive of data: Every form containing 120 compound characters, and having a total of 2500 forms. So there are contain extensive data that are pre-processing raw data into image data and also label this data. It was such a tough job for us.
- Optimize the hyperparameter of the deep convolutional neural network.
- High-performance machine for the extensive network.

# CHAPTER 3

# Research Methodology

## 3.1 Introduction

MatriVasha is a Bangla compound character dataset. This dataset can be used in different ways. MatriVasha dataset is consists of 120 unique compound characters. MatriVasha dataset collected from various kinds of students like university students, college students, school students in Bangladesh, and a total of 2600 people fill-up the forms where 50% from the male, and the other 50% from the female. After the checking and processing of the full dataset, we have 2552 people correct data were 1267 male and 1285 female. Initially, the compound characters are written in a form. Then scanned it in JPEG format. After scanning, we have checked all data manually. Then after many stepped, we get a single compound character image.

But before selecting these 120 compound characters, we have researched in this sector. Because in Bangla language have more than 200 compound characters. But from more than 200 characters, some character isn't using. So we don’t need to collect this type of character. For selecting our desire 120 characters, we randomly select some text from Bangla newspapers using a python script. After then, we researched on the text that, which compound characters used frequently. After then, we select the top 120 characters for our MatriVasha dataset.

Our target people was school, college, a university-level student in collecting our handwritten data. We can easily collect correct data for our research from our students.

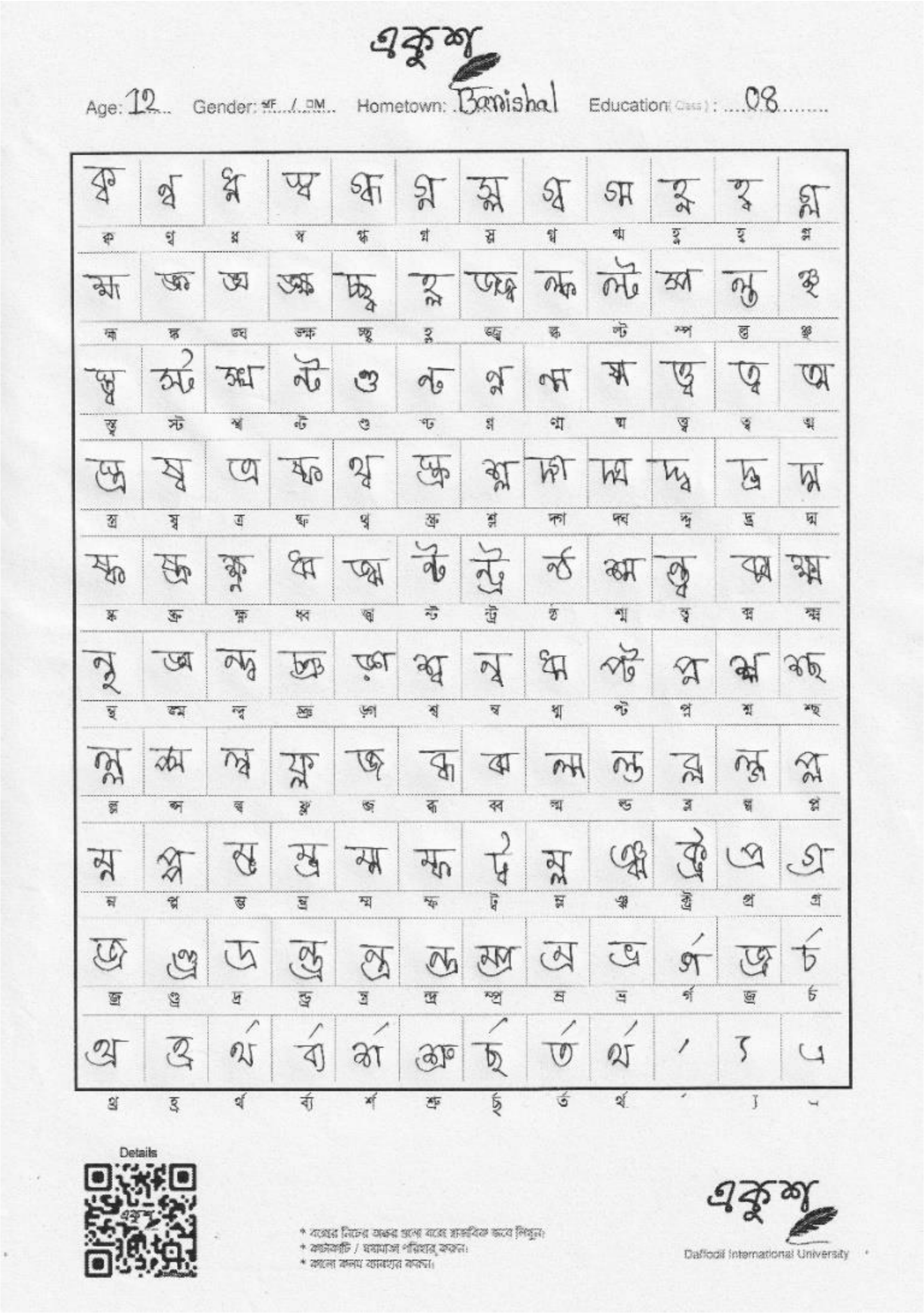

Figure 3.1: Data Collection Form

Figure 3.1 is an example form that we use for data collection.

### 3.2 Data Processing Method

Currently, every researcher is working with machine learning and deep learning. In this area, there are different areas for researches. But for the perspective of our country Bangla language should be the main research area. For this reason, first of all, we have to develop our language for use in machine learning.

First of all, we have to make a perfect dataset for Bangla character recognition., and after some research, we can see there is no perfect dataset for the compound character. In this research, we try to introduce the easiest and fast way to collect data for any language.

Initially, data are collected in the given form. After the collect all data then we are checking those forms manually. Then we separate the correct form as male and female. Then we scanned it in JPEG format with 300dpi. After that, we rename the scanned form. Then we have to crop individual character. For this, we have to follow three-step. Firstly, we cropped our data form as a square which consists of 120 compound characters. Secondly, we crop 120 characters as ten rows, which each row consists of 12 compound character., and lastly, we crop each row., and finally, we get 120 individual characters. Then after some process, we prepare data for character recognition.

Each step we use python script. We can easily, and faster do our work using such a script. For scan our form we used automated G.U.I. in python. Also for cropping image use OpenCV, P.I.L., PyQT. After cropping we transfer our image in different type for multiple purpose. Such as JPEG, grayscale, invert, and CSV.

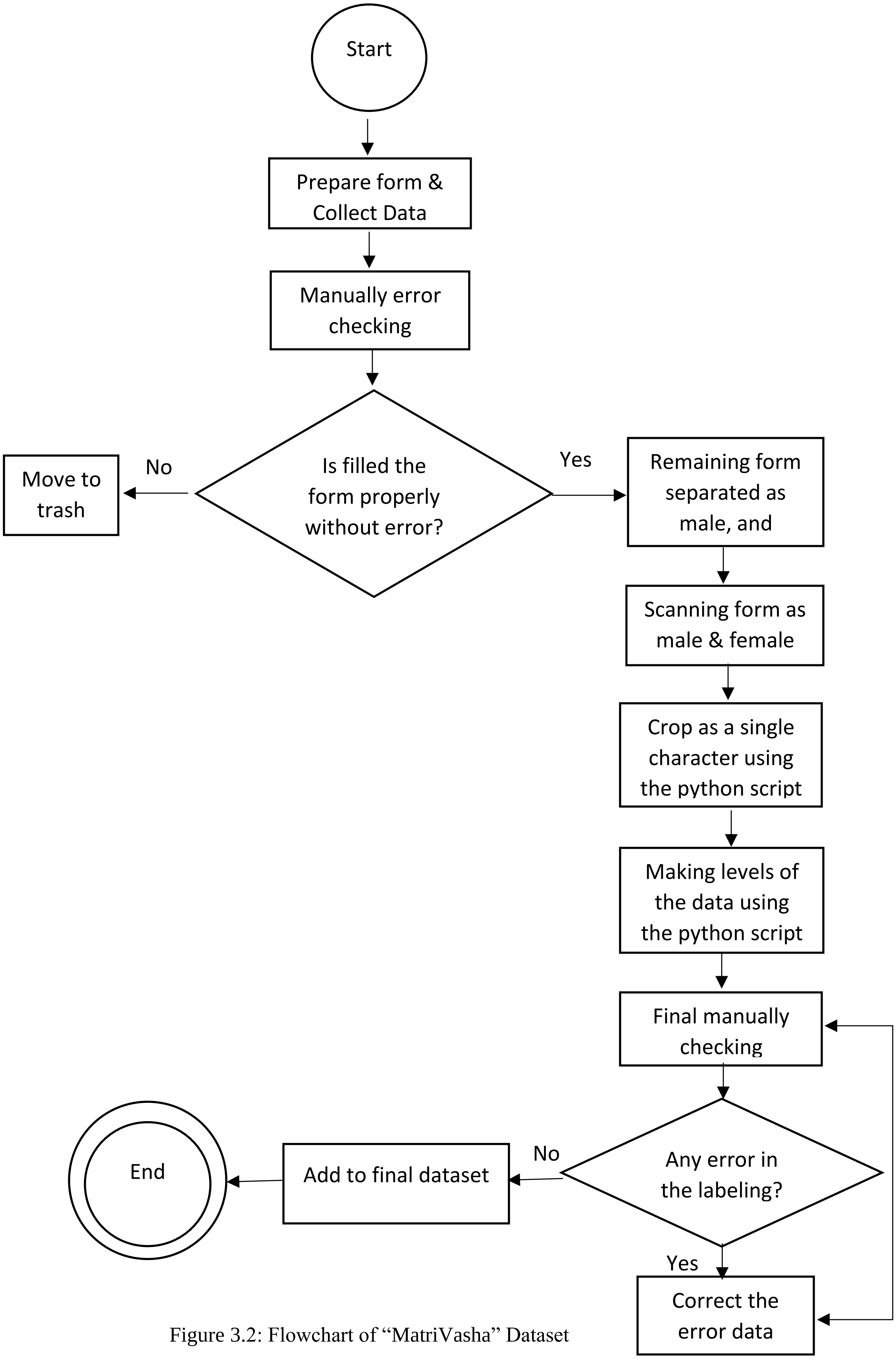

Figure 3.2: Flowchart of “MatriVasha” Dataset

From figure 3.2 flowchart, we try to show the full process for making our dataset. In the below, we tried to describe the full process.

### *3.2.1* Form Processing

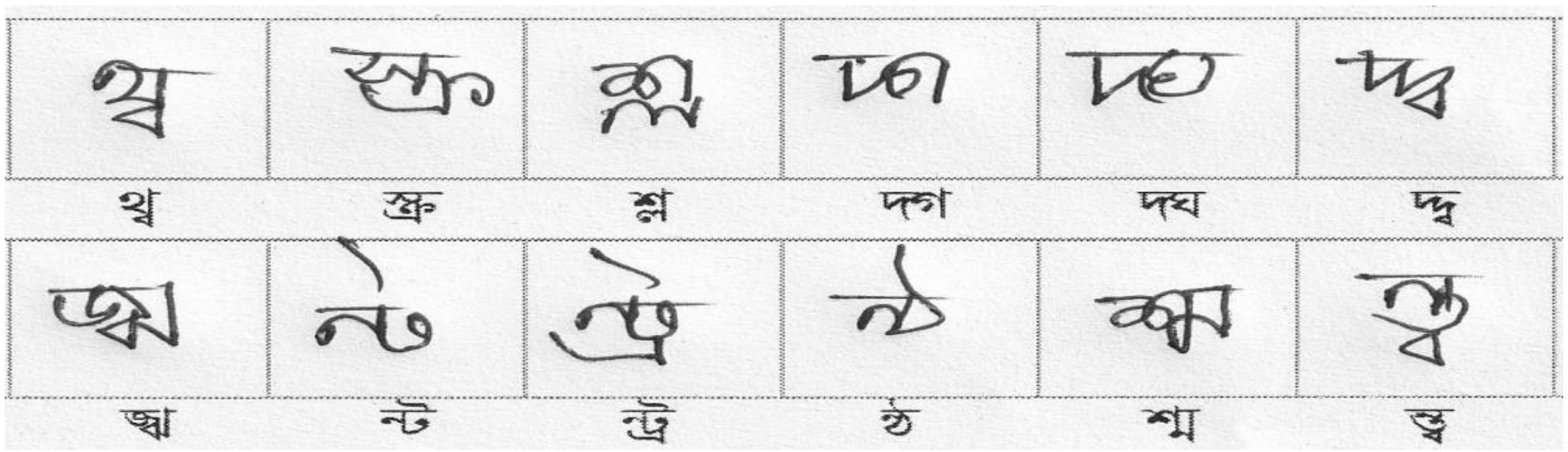

Figure 3.3: Equal size box to fill up the individual character.

There is more than 200 compound character. From those characters, we try to figure out which character is using most frequently. Ekush paper has the 20 most frequent compound characters. Without these 20 characters, we try to select the next 120 most frequent characters. To collect these 120 compound characters, we try to make a perfect form such that after collecting data, we can easily preprocess these data. We have to maintain the same size for collecting individual characters. Otherwise, it's too tough to process these data. We maintain the same square size for all individual boxes. If we follow our form, we can see that. Also, we show the target character under the box. Because in this way, it's easy to write those characters onto the form.

### *3.2.2* Form Scanning

For scanning the form, we use the default app "I.J. Scan Utility" for the Canon scanner. It takes 10sec to scan each form. But it has a big problem that this application cannot repeatedly scan the file automatically. User needs to click the scan button for all image. For this reason, we need more time to scan a form. To solve this problem, we used automated G.U.I. in python. We define a specific position to click the button after a particular time for scanning the file by our python script. Using python script, we need 15 seconds to scan each page with 300dpi in grayscale.

Algorithm 1: Automatically save files after scan.

1. Import necessary python library
2. Find the perfect position for the scan button, and an exit button
3. For all form do a certain number of time:
4. Click scan button using python function
5. Wait for a specific time
6. Click exit button using python function

### *3.2.3* Correction, and Crop

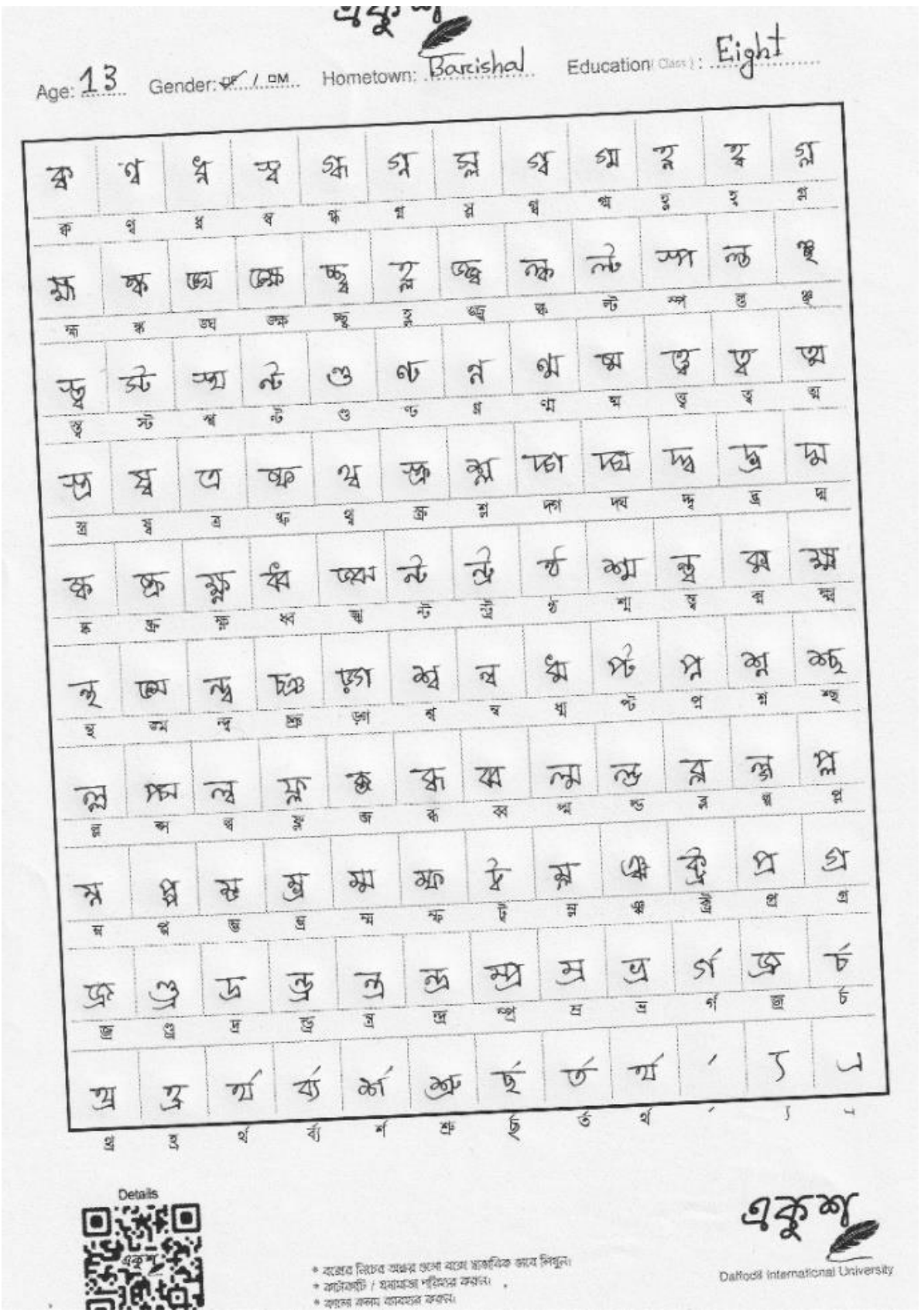

Figure 3.4: Before the cropping

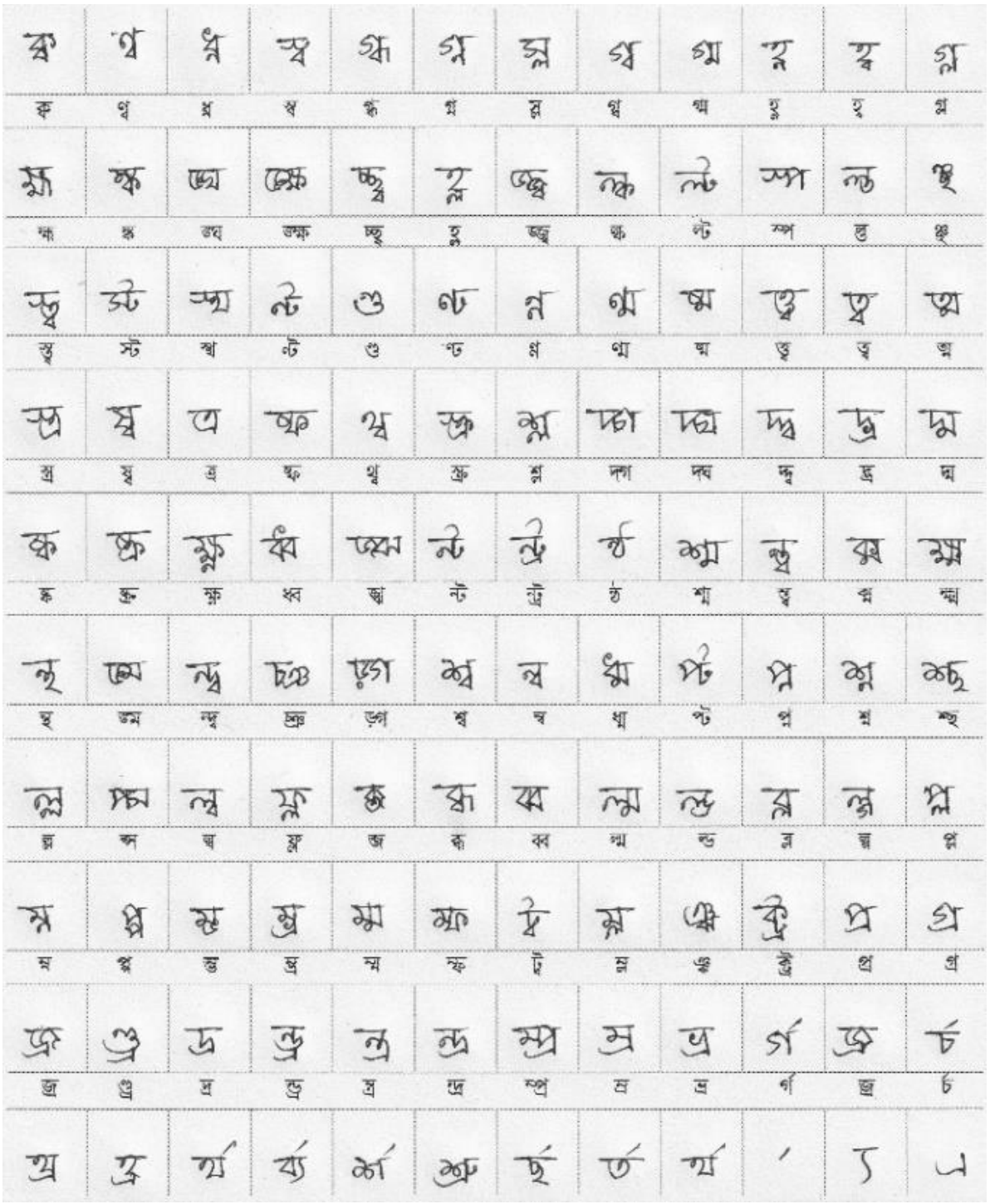

Figure 3.5: After the cropping

In figure 3.4, we can see the original image, and figure 3.5, and we can see the after cropping resize the image. When we scan each form individually then some form changes their angle for some issue. But for our next process, our all form has to be at the same angle. We try to solve our problem using canny edge detection []. This algorithm tries to find the most prominent contour. Our form has a significant black boundary. This algorithm detects this boundary and changes its angle for all forms.

Firstly, we have to find out the detect Canny edge for each image. Then among edge, we have to select a maximum contour area which is actually black boundary. After detecting the area, crop the image., and then save the file.

### *3.2.4* Separate each row

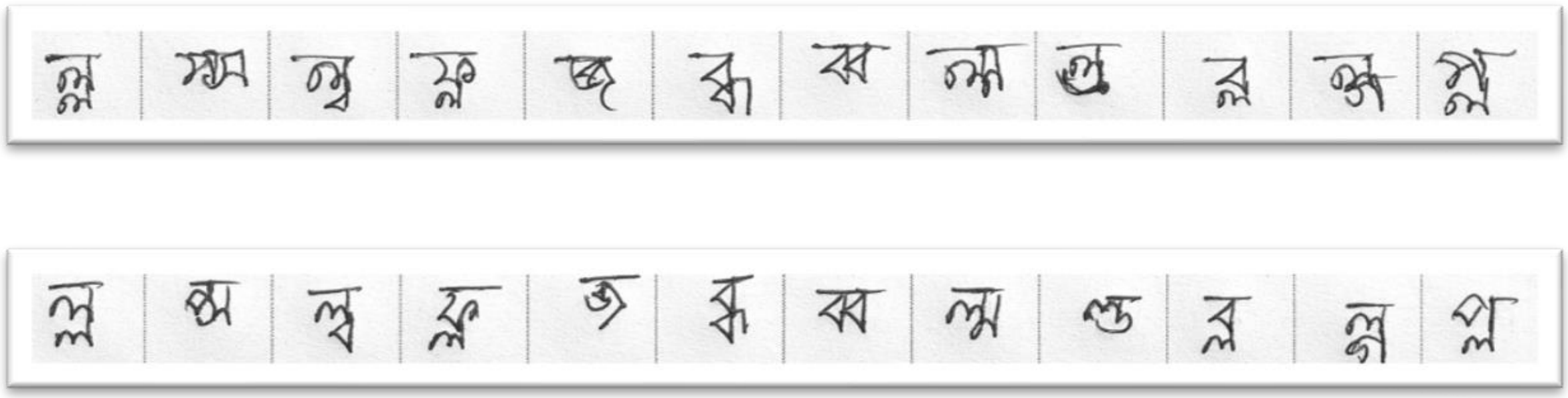

Figure 3.6: row cropping

In figure 3.6, we can see the sample output after row-wise cropping. After the previous step, all our data will be of the same size and same angle. Now we have to separate all characters from the form data. Firstly, we have to crop as row-wise. Then if we crop column-wise from each row we will get individual characters. Each form has 10 row. We save these 10 row in 10 separate folders after each cropping. For each image, we can select the same distance for cropping as row-wise. As when we make our form then we maintain the same distance between each row. We can initialize start and end value for each row. Then each time we can increase the start and end value concerning the distance value. And after cropping each row, we can save the same row in the same file.

#### *3.2.5* Separate each character

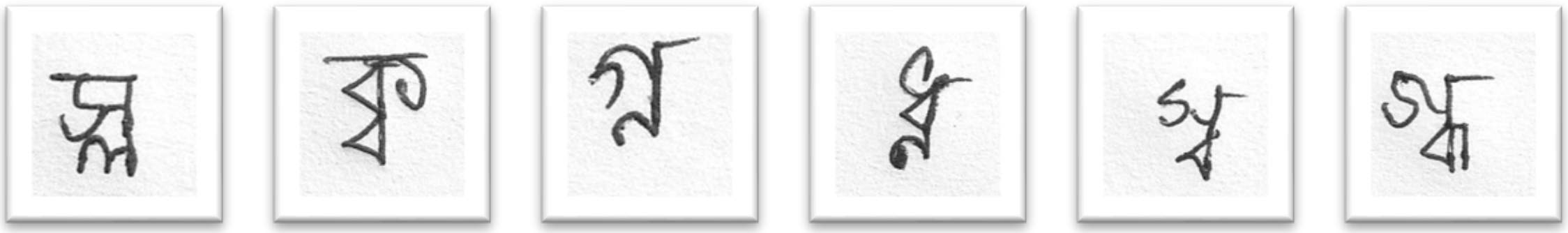

Figure 3.7: Example of after separates each character

In figure 3.7, we can see the example of the output image. After cropping as row-wise, we can see that each row has 12 characters as each small box has equal-sized had so we can maintain the same distance for cropping each box. But now we have to save 120 unique characters in a unique folder. We can easily handle this task in our python script. Now we have to crop each row image as column-wise. We can initial start and end value in the same way. Then crop each square from each row.

In the same way, we can separate individual characters from each row. Each character from the first row we can save 0-11 folder, respectively. Then the second row, we can save 12-23. In the same way, we can save our 120 compound characters.

#### *3.2.6* Noise Remove, and Smooth image

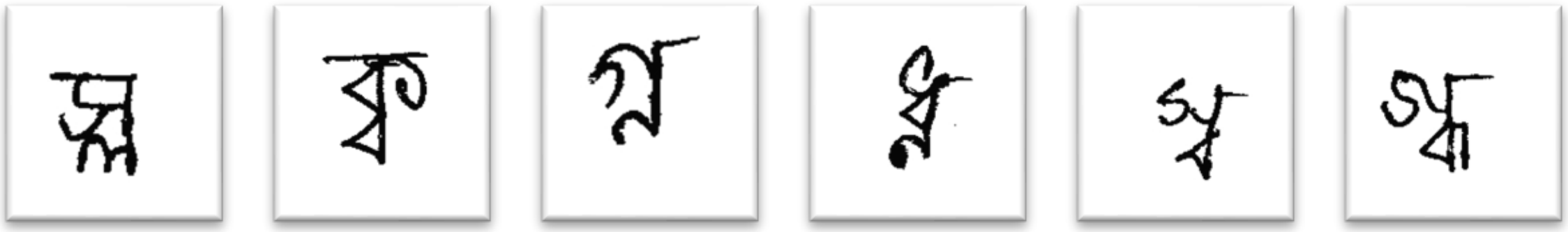

Figure 3.8: Example image after Noise Remove, and Smoothing

The threshold applies fixed-level thresholding to a single-channel array. In the grayscale image, this function used to get a binary level image., and for the filtering image, this function removes small or large noise value from image.  This function works in a two-step. Firstly remove noise from the image and secondly smooth the image. Also, after removing the noise, add a unique value O.S.T.U. [24]., and Gaussian blur used to smooth the image. In figure 3.8, we can see the example of the output image.

### *3.2.7* Removing Unnecessary Information

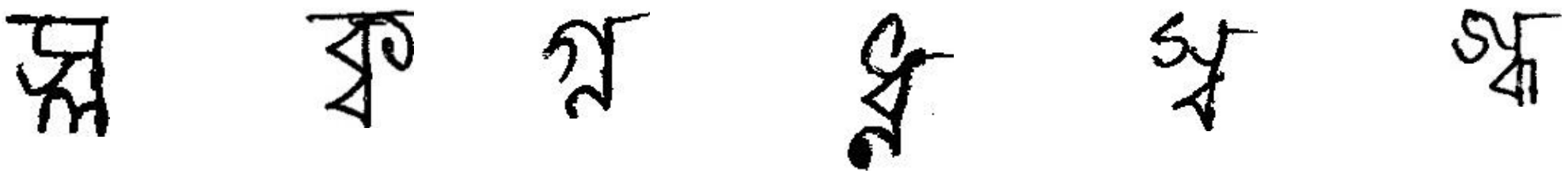

Figure 3.9: Example image after removes unnecessary bit.

For good accuracy in machine learning, we remove unnecessary information from our image. We remove all extra white pixels. Then we get only black pixels for each character.

### *3.2.8* Invert the Images

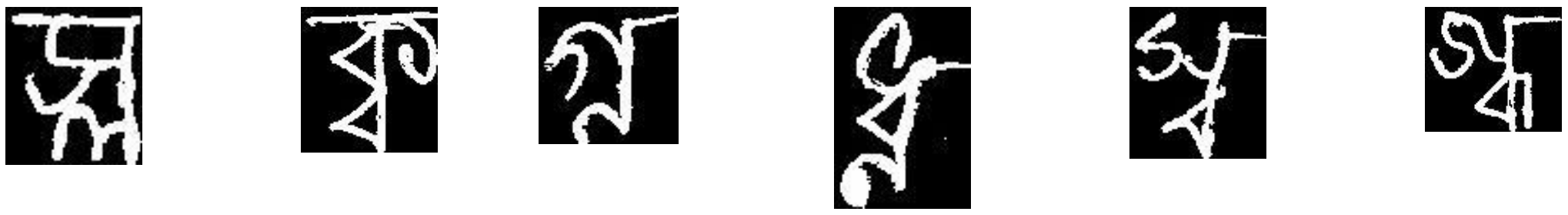

Figure 3.10: Example images after inverting the images.

In this step, we just invert the current image bit. The white bit will be black, and the black bit will be white. That means the white part will be black, and the black part will be white. In figure 3.10, we can see the example of inverted images.

## 3.3 MatriVasha Dataset

MatriVasha the largest dataset of handwritten Bangla compound characters for research on handwritten Bangla compound character recognition. In recent years application-based researchers in machine learning and deep learning have achieved interest, and the most significant is handwriting recognition. Also, Bangla O.C.R. has a tremendous application. Also, then the Bangla alphabet is the fifth most popular composition method in the world. Therefore, we are tries to introduced Bangla handwritten compound character dataset. The proposed dataset contains 120 different types of compound characters that consist of 30624 images written by 2552 people, where 1267 male, and 1285 female in Bangladesh. This dataset can be used for other issues such as gender, age, district base handwriting research because the sample was collected that included district authenticity, age group, and an equal number of men and women. It is designed to create an accreditation system for handwritten Bangla compound characters., and this dataset is freely available for any kind of academic research work.

### *3.3.1* Constructing MatriVasha

MatriVasha dataset of Bangla handwritten compound characters was collected from 2552 people were 1267 male and 1285 female in Bangladesh. MatriVasha dataset contains 120 different types of compound characters that consist of 306240 images. After preprocessing the dataset, Bangla's handwritten compound characters are confusing, and some are similar. So, when writing the people, they made the mistake of understanding which character is actually in the form. Then we start to check manually and found 5 compound characters. Most of the people were a mistake to understand the proper compound characters. These authors were selected from a different age, gender, and educational background groups. Figure 3.12 showing a bar chart of data samples from different ages is based on gender. The collected datasets pre-processed in three different formats:

The foreground and the background are reversed so that the image has a black background painted with white color.

The foreground and background are white so that the images black tint with white. Sound and smoothing attempts have been removed using trash holding, and Gaussian blur filters. The advanced dataset is further filtered, and the divergent format is created with the CSV after the required smoothing.

Table 3.1: Number of character in MatriVasha by gender

| Gender | Compound Characters | Total In MatriVasha |
|---|---|---|
| Female | 1285 | 2552 |
| Male | 1267 | |

The MatriVasha will be available in a variety of formats, depending on the user's desired applications, as well as without additional information on character images, and will save on the aspect ratio of the padding, and be added to the CSV format. Table 3.1 showing the details of the MatriVasha dataset base on gender.

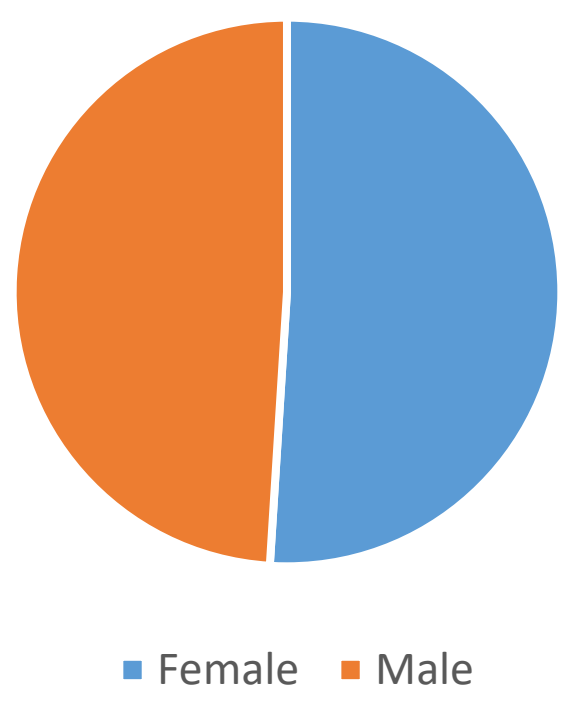

Figure 3.11: A Pie Chart for comparison of Male, and Female Data.

In figure 3.11 showing a pie chart of databases on gender. Our dataset has almost 50% male data, and almost 50% of female data.

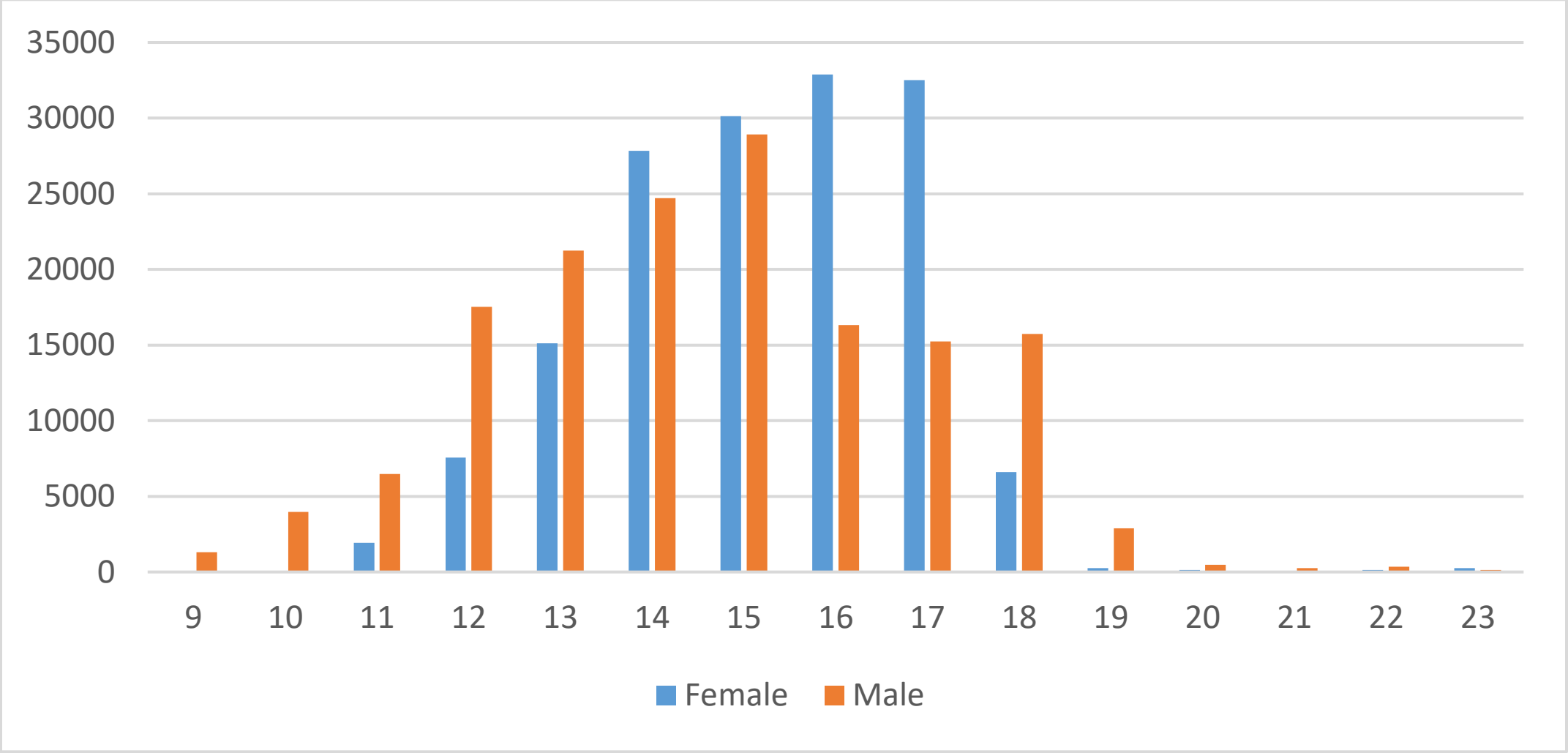

Figure 3.12: A Bar Chart of different age's student base on gender

In figure 3.12, showing a bar chart of data samples from different ages based on gender.

### *3.3.2* Visual Representation of MatriVasha Dataset

MatriVasha dataset contains 120 different types of compound characters that consist of 30624 images. In Figure 3.13, Figure 3.14, and Figure 3.15, we try to see the number of single images in our dataset.

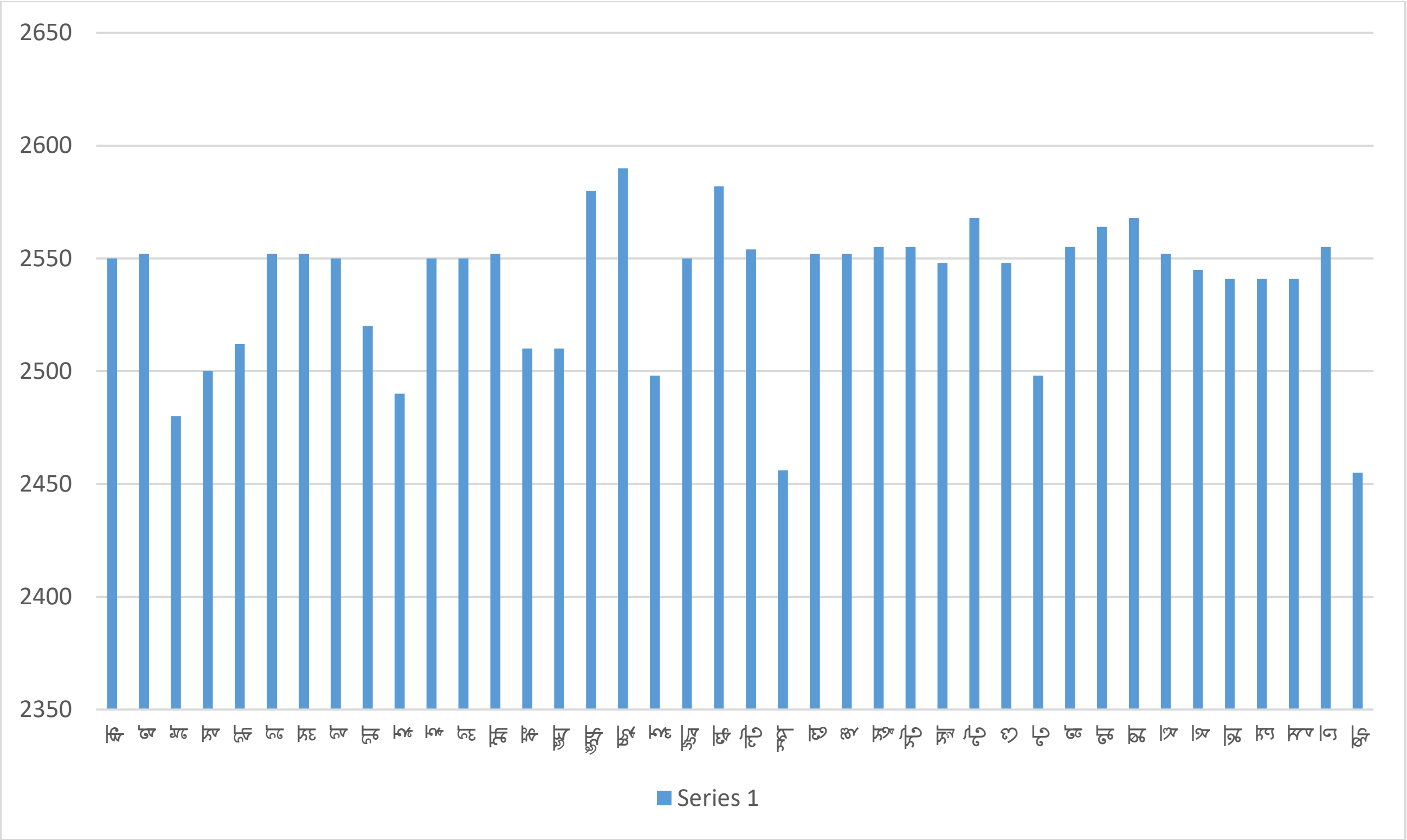

Figure 3.13: First 40 Compound Character Frequencies in MatriVasha Dataset.

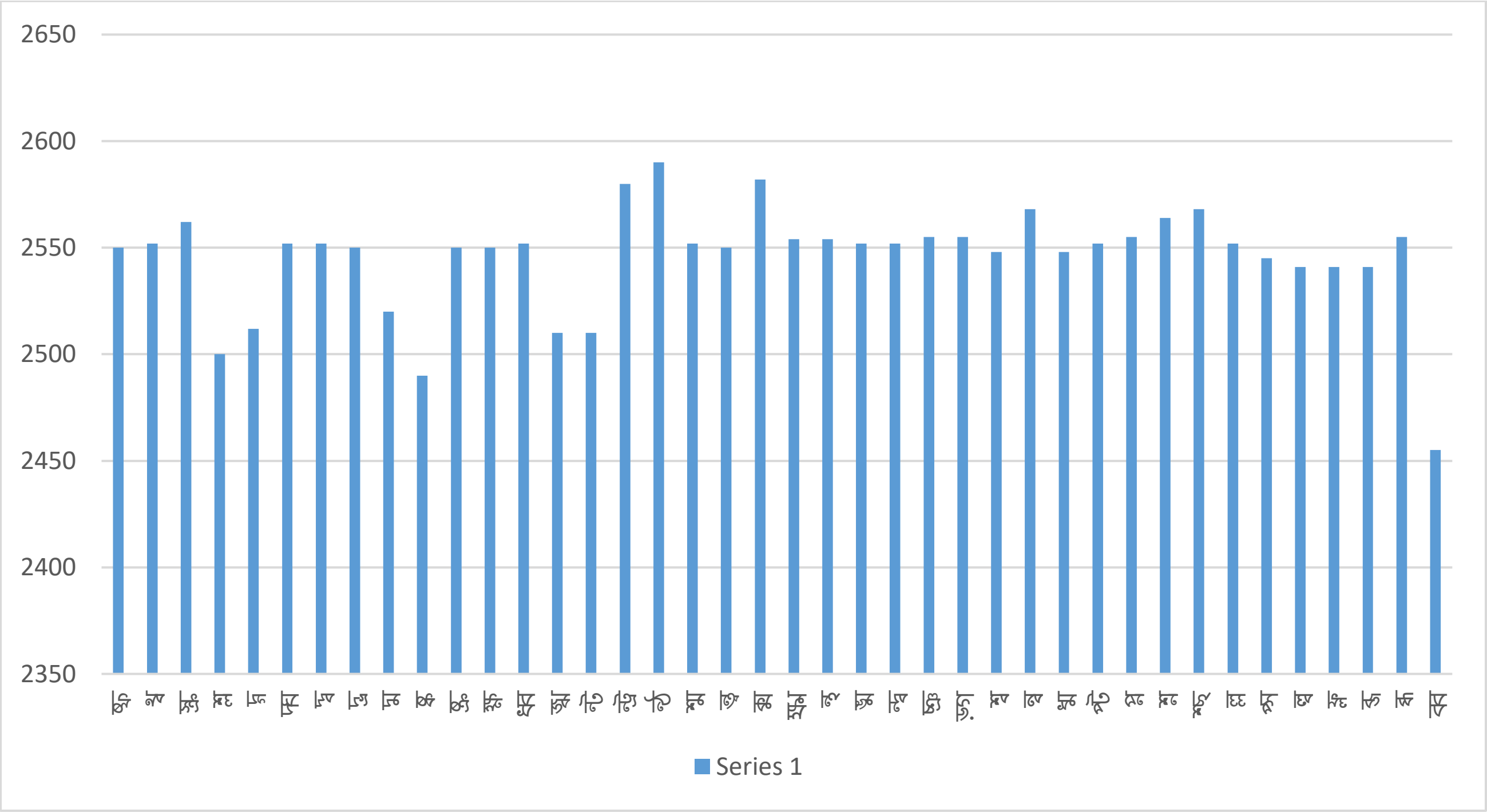

Figure 3.14: Second 40 Compound Character Frequencies in MatriVasha Dataset.

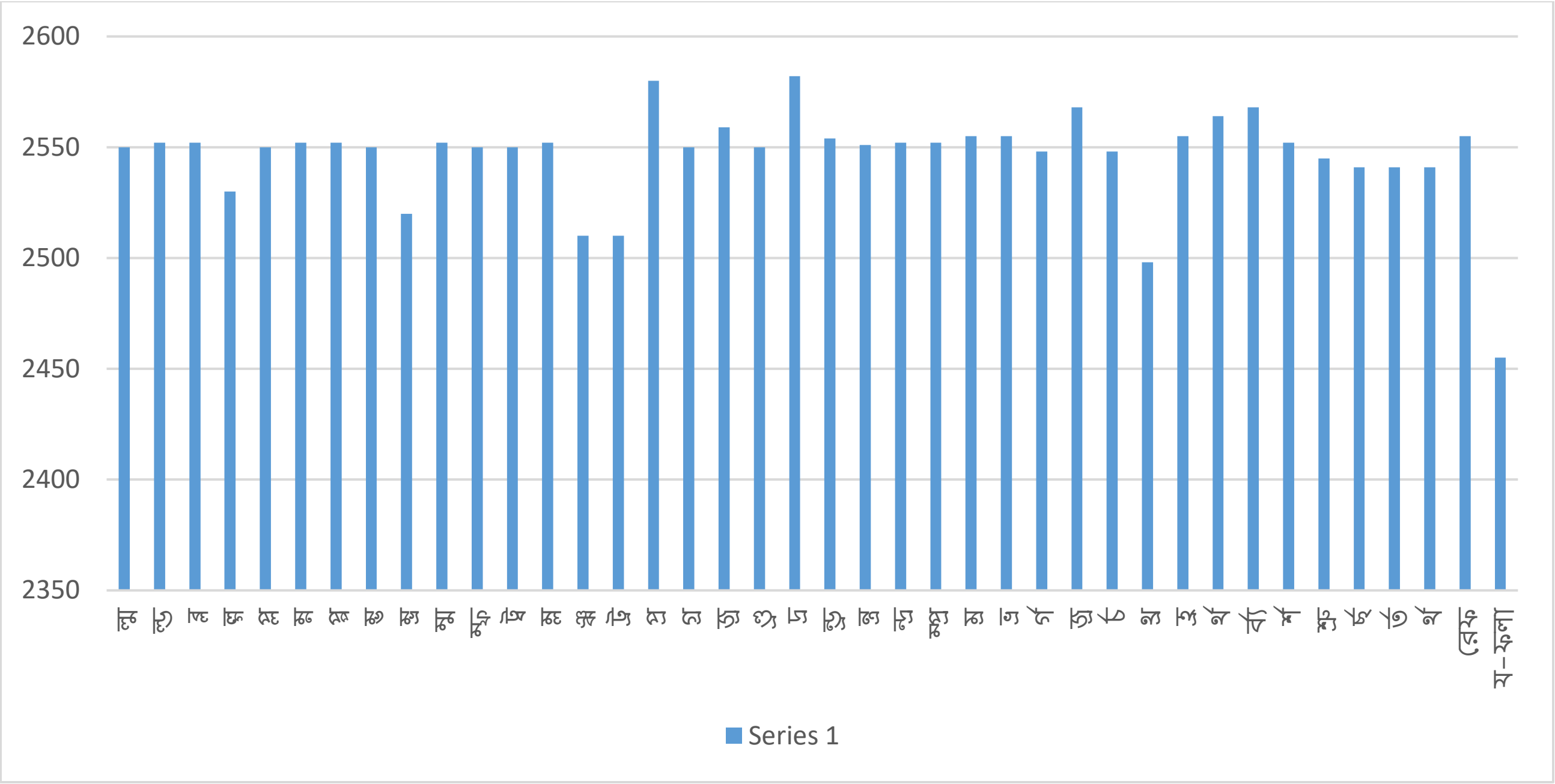

Figure 3.15: Last 40 Compound Character Frequencies in MatriVasha Dataset.

### *3.3.2* Data Labeling

The image is stored as a JPG format, and each image has a unique I.D. or name that represents the people's gender, hometown, age, education level, and serial number. Data labeling also storing people information that identifies the person who uses the id. So that this dataset not only uses handwriting recognition but also helps to predict a person's gender, age, and location, as well as help investigators, focus more on a specific category for suspicious and forensic purposes. This id or name is set according to the following criteria, with the first one-digit indicating the gender of the author. Here the number identifies gender and education. If the number is 0, that means the writer was male, and the number is one that means the writer was female. The next one represents hometown names, here 3 or 4 characters represent hometown. Then next come to age, here fill up age which is given in the form. Then they give their education or occupation level using the number to understand which level they belong. Here, 0 means the primary level (0-5), one means high school level (6-10), two means college level (11-12), three means university, and the last one is the serial number of each form, and it's separated using underscore (_). Here an example

1_DHA_25_3_00032

So, here the first character represents it was written by the female who is from Dhaka district, and her age is 25, and she is a university student, and the last one belonging to the serial number of this data.

### *3.3.3* Possible Uses of MatriVasha Dataset

MatriVasha dataset can be used to the recognized compound character, and also use to some research fields like Bangla O.C.R., Deep learning, and machine learning. On the other hand, the prediction of age, gender, the location from handwriting is a very interesting research area. This information can even be used for forensic purposes, where it can help investigators focus more on suspects in a particular category. Nowadays, many people are not famous in a compound character, MatriVasha can help people to know many unknown compound characters, and its help to make a meaningful sentence.

## 3.4 MatriVasha

MatriVasha is the largest dataset of handwritten Bangla compound characters for research on handwritten Bangla compound character recognition. MatriVasha belongs to different types of compound characters which are written by the students. In recent years application-based researchers in machine learning and deep learning have achieved interest, and the most significant is handwriting recognition. Researchers are more interested in different research types of problems. Also, Bangla O.C.R. has a tremendous application. Also, then the Bangla alphabet is the fifth most popular composition method in the world. In Bangla language there are belongs to basic character, numerical number, compound characters. But many people are not identifying many compound characters. Therefore, we are tries to introduced Bangla handwritten compound character dataset which helps to people to identifying compound characters. The proposed dataset contains 120 different types of compound characters that consist of 306240 images written by 2552; people were 1267 male and 1285 female in Bangladesh. This dataset can be used for other issues such as gender, age, district base handwriting research because the sample was collected that included district authenticity, age group, and an equal number of men and women. People take it seriously, and they write these compound characters for enhanced their knowledge, and they fill up form very sincerely. It is designed to create an accreditation system for handwritten Bangla compound characters., and this dataset is freely available for any kind of academic research work that helps to their research and also helps to enhance their knowledge. MatriVasha is a path to introduce the different types of compound characters.

### *3.4.1* MatriVasha Architect

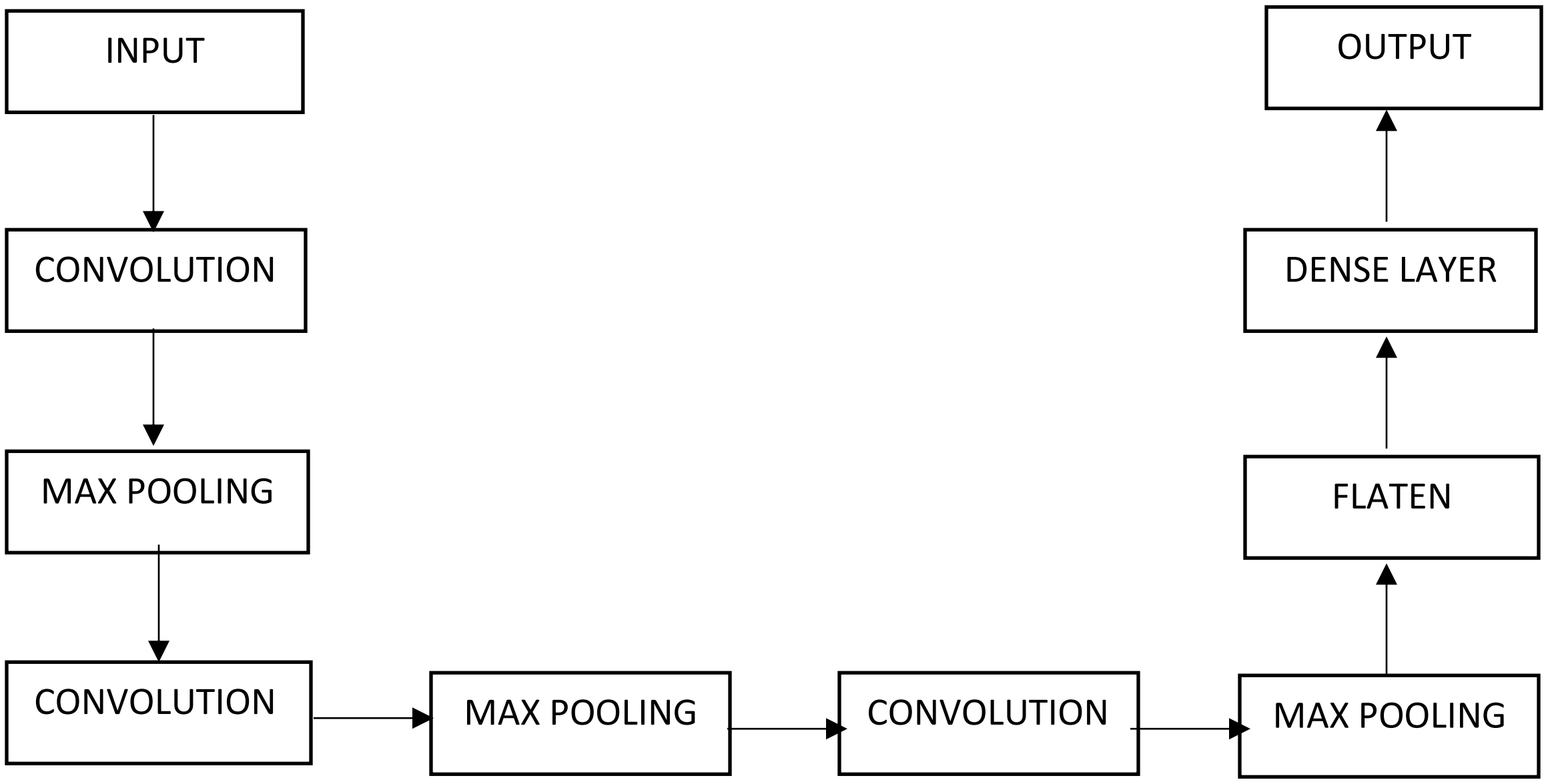

Figure 3.16: Architecture of MatriVasha

For classifying Bangla Handwritten Compound Characters, our proposed model used CNN with Keras. Our model used Convolution, max-pooling layer, flatten, dense layer. We use three convolutional layers. In Layer 1, we have a 32 size filter and three size kernel. In Layer 1, we have a 64 size filter, and three size kernel. In Layer 1, we have a 128 size filter and three size kernel.

Also, there, and 2*2 size three max polling layers. In Keras, we can stack up layers one by one by adding our target layers. We add a convolutional layer with Conv2D (). As we are working with the image, so we use this function. In figure 3.16, we show our proposed model.

| Layer (type) | Output Shape | Param # |
|---|---|---|
| conv2d_53 (Conv2D) | (None, 28, 28, 32) | 896 |
| leaky_re_lu_26 (LeakyReLU) | (None, 28, 28, 32) | 0 |
| max_pooling2d_43 (MaxPooling | (None, 14, 14, 32) | 0 |
| conv2d_54 (Conv2D) | (None, 14, 14, 64) | 18496 |
| leaky_re_lu_27 (LeakyReLU) | (None, 14, 14, 64) | 0 |
| max_pooling2d_44 (MaxPooling | (None, 7, 7, 64) | 0 |
| conv2d_55 (Conv2D) | (None, 7, 7, 128) | 73856 |
| leaky_re_lu_28 (LeakyReLU) | (None, 7, 7, 128) | 0 |
| max_pooling2d_45 (MaxPooling | (None, 4, 4, 128) | 0 |
| flatten_15 (Flatten) | (None, 2048) | 0 |
| dense_29 (Dense) | (None, 128) | 262272 |
| leaky_re_lu_29 (LeakyReLU) | (None, 128) | 0 |
| dense_30 (Dense) | (None, 10) | 1290 |

Total params: 356,810
Trainable params: 356,810
Non-trainable params: 0

Figure 3.17: MatriVasha model architect summary

In figure 3.17, we see the summary of our MatriVasha model architect. After developing our model, we train this model in our train data. And test our model by our test data. After using various kinds of the optimizer, we compile our model. Also, we have to change the model hyperparameter. We use binary cross-entropy for our loss function.

### 3.5 Implementation Requirements

For finishing the above work, such as a statistical report, python script, to develop the model, we need some hardware and tools. When a researcher works in Bangla language, as usual, they need this kind of tool.

## Hardware/Software Requirements:

- Any Operating system.
- GPU
- Minimum Hard disk 10 G.B.
- Minimum RAM 4 G.B.
- Web browser

**Developing Tools:**

- Python 3.7
- Jupyter Notebook
- Sklearn
- Pyqt
- Open CV
- Keras
- Tensorflow

# CHAPTER 4

# Experimental Results, and Discussion

## 4.1 Introduction

Currently, MatriVasha is the largest dataset for compound characters. MatriVasha dataset has 2552 unique people correct data were 1267 male data, and 1285 female data.

We split our dataset for train and test data. The training data is used to train our model with unknown data., and then, for testing our model, we used another unknown data. For better performance, we tune our model hyper-parameters.

## 4.2 Sample Dataset

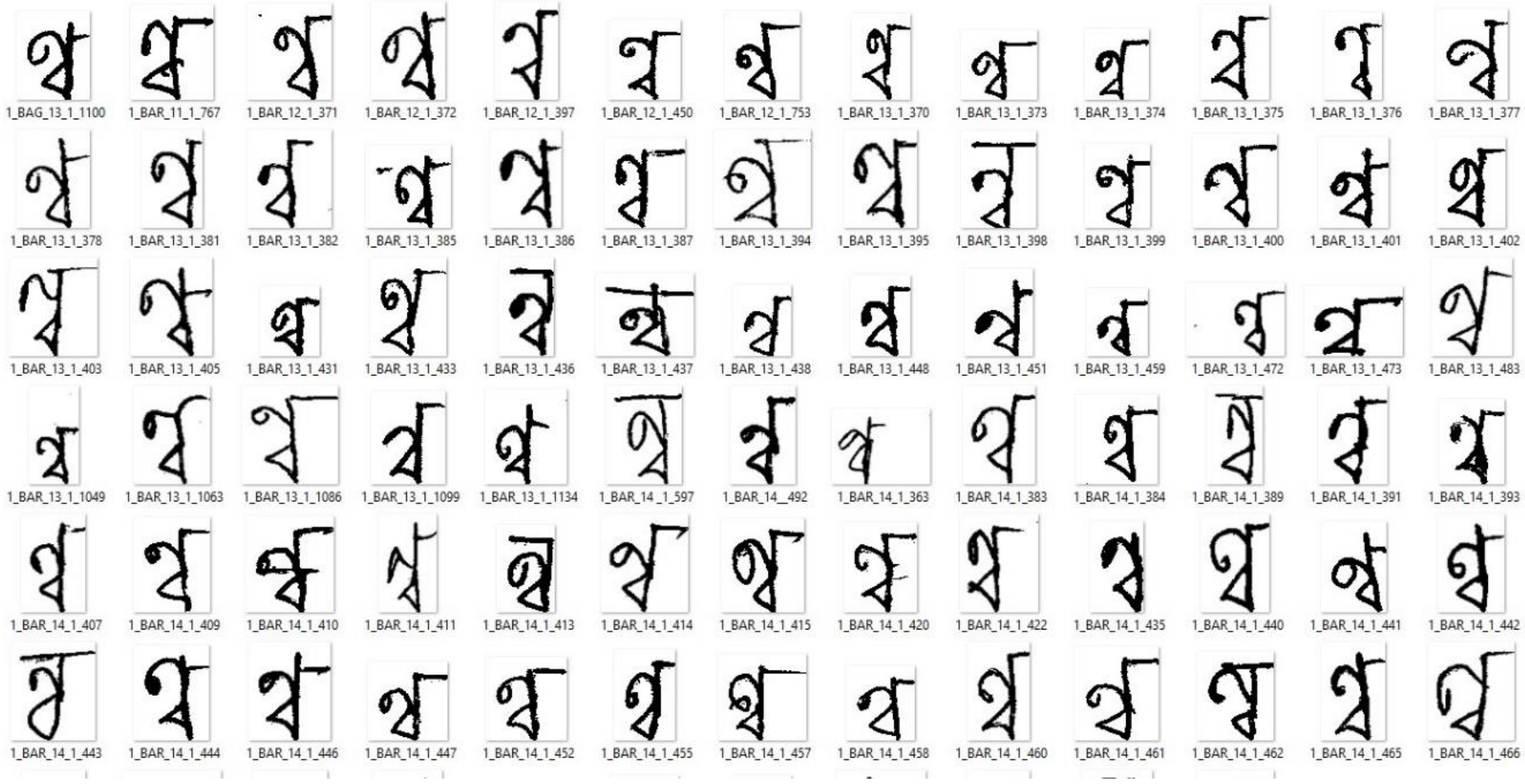

Figure 4.1: Sample Dataset

MatriVasha dataset consists of 2552 unique people correct data were 1267 male, and 1285 female. We built our model such that our model train, and test by image data. So we store our data as images. In Figure 4.1, we show the sample dataset. In this way, we store our 120 unique compound characters' images., and our dataset has a total of 306,240 images.

## 4.3 Experimental Results

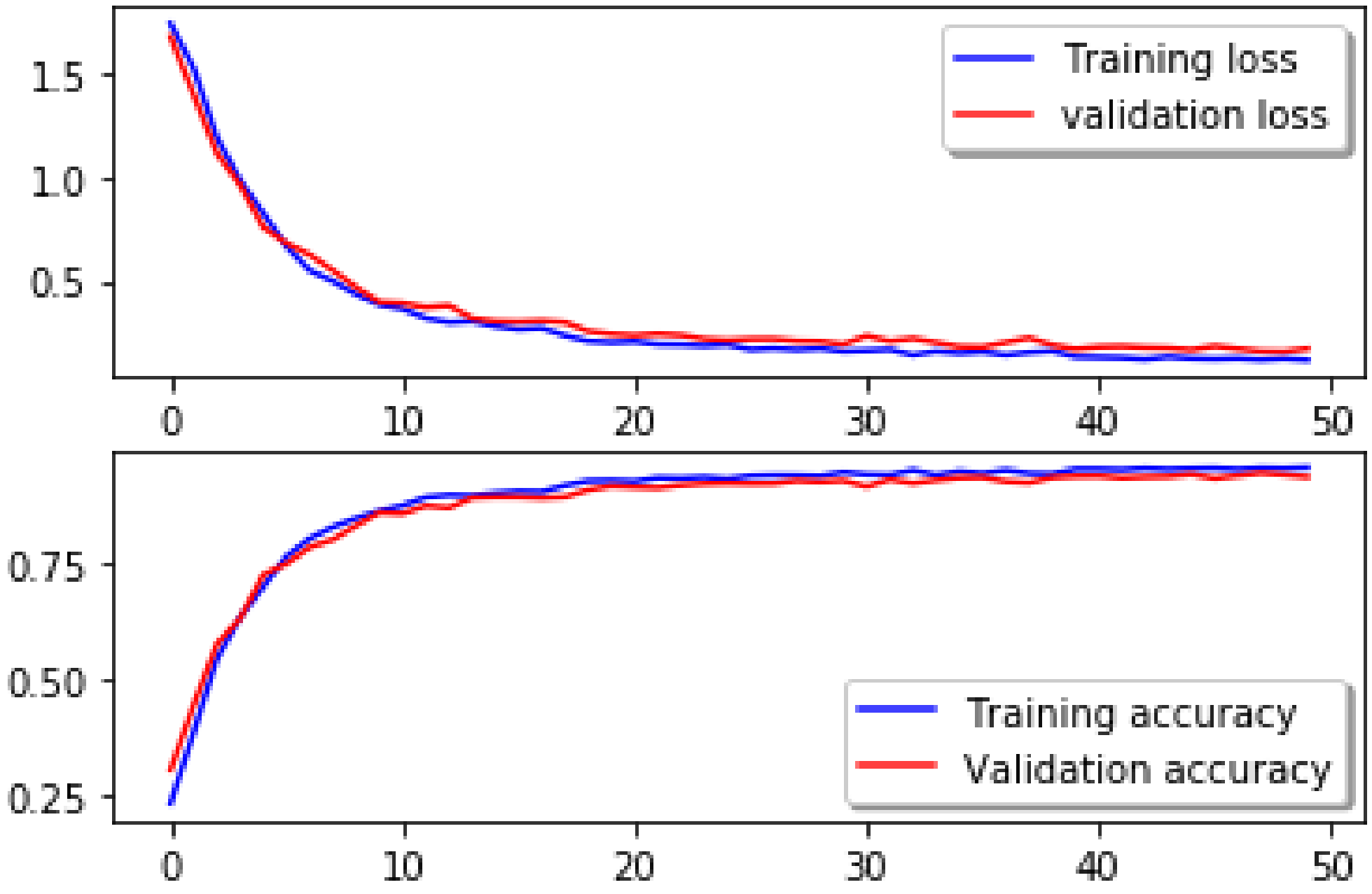

Figure 4.2: Accuracy and loss

We got 94.49% accuracy from our model. Because all our data is not finished checking. Now we are checking 306,240 data manually because our data has some error. We are pretty much sure that after finishes our checking, the accuracy rate will be increased. In Figure 4.1, we can see the training and validation accuracy. And in Figure 4.2, we can see the training and validation loss.

# CHAPTER 5

## Summary, Conclusion, Recommendation, and Implication for Future Research

### 5.1 Summary of the Body

This study introduces the Bangla handwritten compound character datasets. It is possible to develop a compound character recognition system for any database. By applying this proposal method, researchers will benefit from various research areas. They will also help in the automatic recognition of some features of the author, such as age, gender, location. For any purpose of research, forensic databases are essential elements in which data processing plays an important role. Also, provide a model to a dataset MatriVasha in recognition of Bangla compound character. The recognition rate found by the system was 94.49% of the MatriVasha dataset.

### 5.2 Conclusion

This research initially led to the creation of a diverse repository for computer vision and N.L.P. research, and the dataset was known as MatriVasha. Initial tests using a sophisticated identifier demonstrate the challenge of recognition. It leaves a great place for improvement and encourages the community to seek new accreditation methods. It represents a new and modern performance standard for the current generation of classification and learning systems.

To understand the model, we understand that the convolutional neural network can achieve better performance in classifying and recognizing compound characters. CNN does an excellent job of recognizing compound character for its distinctive features. A sizeable compound character dataset that helps to enhance the robustness of this approach for compound character recognition of Bangla handwriting.

### 5.3 Recommendations

As a high configuration computer, the GPU would have been better than the result of this model. Also, using an efficient dataset can produce a better output of this research work.

## 5.4 Implication for Further Study

In the future, we will extend our dataset with more compound characters. More accuracy can be achieved by providing a better collection of datasets. Also, we'll create a website where users can download the form, and upload a scanned copy, which will automatically process those data, and add to the dataset after verification. This website allows users to search and download character data by age, gender, and districts.